\documentclass[preprint,authoryear]{elsarticle}

\usepackage{etoolbox}
\usepackage{expl3}
\usepackage{framed,multirow}
\usepackage{booktabs}       
\usepackage{amsmath,amssymb,amsfonts}
\usepackage{nicefrac}
\usepackage[ruled,linesnumbered]{algorithm2e}
\usepackage{graphicx}
\usepackage{subfig}
\usepackage{tikz}
\usetikzlibrary{shapes,arrows}
\usepackage[toc,page]{appendix}
\usepackage{url}
\usepackage{hyperref}

\graphicspath{{Figs/}}

\def\x{{\mathbf x}}

\DeclareMathOperator*{\argmax}{arg\,max}
\DeclareMathOperator*{\argmin}{arg\,min}
\DeclareMathOperator{\E}{\mathbb{E}}

\makeatletter
\def\ps@pprintTitle{%
	\let\@oddhead\@empty
	\let\@evenhead\@empty
	\def\@oddfoot{\footnotesize\itshape
		{To appear in Computerized Medical Imaging and Graphics, DOI: \url{https://doi.org/10.1016/j.compmedimag.2022.102087}}}
	\let\@evenfoot\@oddfoot
}
\makeatother

\begin{document}

\begin{frontmatter}

\title{NPBDREG: Uncertainty Assessment in Diffeomorphic Brain MRI Registration using a Non-parametric Bayesian Deep-Learning Based Approach}
\author[1]{Samah Khawaled\corref{cor1}}
\cortext[cor1]{Corresponding author}
\ead{ssamahkh@campus.technion.ac.il}
\author[2]{Moti Freiman \corref{cor2}}
\cortext[cor2]{is a Taub fellow (supported by
the Taub Family Foundation, Technion’s program for leaders in Science and Technology).}
\ead{moti.freiman@technion.ac.il}

\address[1]{Department of Applied Mathematics, Technion – Israel Institute of Technology, Haifa, Israel}
\address[2]{Faculty of Biomedical Engineering, Technion – Israel Institute of Technology, Haifa, Israel}

\begin{abstract}
	Quantification of uncertainty in deep-neural-networks (DNN) based image registration algorithms plays a critical role in the deployment of image registration algorithms for clinical applications such as surgical planning, intraoperative guidance, and longitudinal monitoring of disease progression or treatment efficacy as well as in research-oriented processing pipelines. Currently available approaches for uncertainty estimation in DNN-based image registration algorithms may result in sub-optimal clinical decision making due to potentially inaccurate estimation of the uncertainty of the registration stems for the assumed parametric distribution of the registration latent space.   
	We introduce NPBDREG, a fully non-parametric Bayesian framework for uncertainty estimation in DNN-based deformable image registration by  combining an \texttt{Adam} optimizer with stochastic gradient Langevin dynamics (SGLD) to characterize the underlying posterior distribution through posterior sampling. Thus, it has the potential to provide uncertainty estimates that are highly correlated with the presence of out of distribution data. We demonstrated the added-value of NPBDREG, compared to the baseline probabilistic \texttt{VoxelMorph} model (PrVXM), on brain MRI image registration using $390$ image pairs from four publicly available databases: MGH10, CMUC12, ISBR18 and LPBA40. The NPBDREG shows  a better correlation of the predicted uncertainty with out-of-distribution data ($r>0.95$ vs. $r<0.5$) as well as a $\sim7.3\%$  improvement in the registration accuracy (Dice score, $0.74$ vs. $0.69$, $p \ll 0.01$), and a $\sim18\%$ improvement in registration smoothness  (percentage of folds in the deformation field, 0.014 vs. 0.017, $p \ll 0.01$).
	Finally, NPBDREG demonstrated a better generalization capability for data corrupted by a mixed structure noise (Dice score of $0.73$ vs. $0.69$, $p \ll 0.01$) compared to the baseline PrVXM approach. 
\end{abstract}
\begin{keyword}
 Brain MRI\sep Bayesian Deep-learning
\sep   Deformable image registration \sep Uncertainty estimation 
\end{keyword}

\end{frontmatter}

\section{Introduction}

Deformable image registration is a task fundamental to a wide-range of image-guided clinical applications, including: image-based surgical planning and intra-operative guidance \cite{luo2019applicability}, motion compensation, \textit{inter} or \textit{intra} subject alignment for change detection, and longitudinal analysis, among others \cite{hill2001medical,zitova2003image}.

Estimation of the uncertainty in the registration algorithm results plays a vital role in balanced clinical decision decision-making. For example, it allows surgeons to evaluate the risk involved in image-guided tumor resection by quantifying the reliability of the registered image data. Inaccurately calculated registration may results in severe implications to the patient due to over or under estimated safety margins for the resection \cite{luo2019applicability,jackson2021effect}. Accurate estimation of the uncertainty in the registration will enable surgeons to reliably assess the proper safety margins for the resection.

In light of the success of DNN-based methods in numerous computer vision tasks, several studies aimed to propose deformable registration approaches based on DNN models that are more efficient and less time-consuming compared to their classical counterparts \cite{dalca2019learning,balakrishnan2018unsupervised,dalca2018varreg,yang2017quicksilver,shao2021prosregnet,de2019deep}. These techniques learn a deformation field prediction model either through a supervised learning framework (i.e with the help of provided reference deformation fields) \cite{yang2017quicksilver} or in an unsupervised manner \cite{balakrishnan2018unsupervised,dalca2018varreg,de2019deep}. 

\begin{figure*}[t!]
    \centering
    \includegraphics[height=0.2\linewidth]{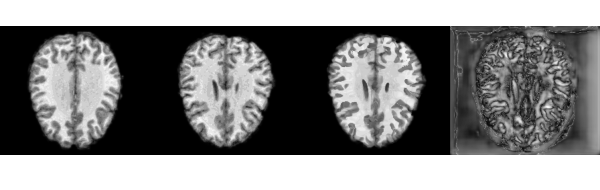} \\
    \includegraphics[height=0.2\linewidth]{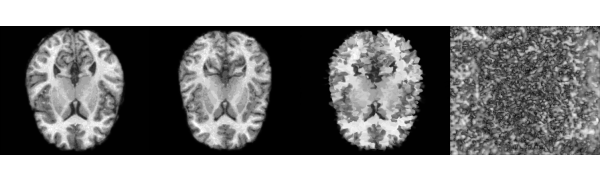}
    \caption{Importance of uncertainty assessment in registration of brain MRI image to a given template. From left to right: Brain MRI image, fixed template, the resulting warped image and the voxel-wise uncertainty map. The first row presents an example in which the DNN predicted a deformation field that successfully mapped the patient data onto the template data while the second row presents a failure of the DNN to predict the deformation field. } 
    \label{fig:introexample}
\end{figure*}

However, the practical utilization of DNN-based image registration methods in clinical applications is hampered by the lack of computational mechanisms for quantifying the uncertainty in the registration predicted by the model. For example, Fig.~\ref{fig:introexample} depicts two cases of success and failure of the network to predict the correct deformation field without a noticeable difference in the input images. 

Bayesian DNN-based models have the potential to enable a safer utilization of DNN approaches for medical image registration, improve generalization capacity, and to provide uncertainty measures of the predictions by characterizing the entire posterior distribution of the network parameters.

Recently, two main methods were proposed to assess the uncertainty in DNN-based methods for medical image registration. Yang et al. \cite{yang2017quicksilver} estimated the uncertainty of the image registration prediction by their QuickSilver model by repeating the inference for a specific data sample multiple times while maintaining stochastic dropout layers. At each repetition, the dropout layers randomly set some of the network units to $0$. The voxel-wise variance of the outputs then used as an estimate of the uncertainty in the predictions. From a practical perspective this method is simple and efficient, yet, it lacks flexibility since it can be used only on DNN-based methods that trained with dropout layers and cannot be generalized to any DNN architecture. Further, from a theoretical perspective, it may not fully express a meaningful uncertainty estimate that is associated with the network predictions \cite{jospin2020hands,osband2016deep}. 
Dalca et al. proposed a probabilistic \texttt{VoxelMorph} DNN for image registration \cite{dalca2018varreg,dalca2019varreg}, where the uncertainty estimates are obtained through a variational encoder-decoder model. This approach relies on a probabilistic generative model, which computes an approximate posterior distribution, $q_\theta(z|X)$, where $z$ is a velocity field generated by the encoder-decoder model. $q$ is constrained by the prior distribution, $p(z)$, which is assumed to be a normal distribution. The modeling of $z$  as a Gaussian distribution is done by adding a Kullback-Leibler (KL) divergence term, $ KL(p||q)$, to the loss function used to train the DNN.
This approach, however, is limited to specific DNN architectures. Further, it assumes a parametric distribution of the latent space in the form of a Gaussian distribution, which may represent an oversimplification of the unknown true underlying distribution. 

In a shift from previous approaches, we propose to estimate the uncertainty in DNN-based Diffeomorphic deformable image registration with a new, non-parametric, Bayesian approach (NPBDREG). We adopt the strategy of stochastic gradient Langevin dynamics (SGLD) \cite{welling2011bayesian} to sample from the posterior distribution of the network weights during the training process \cite{cheng2019bayesian}. Then, We estimate the posterior distribution of the deformation field during inference by averaging velocity field predictions obtained by the model with the saved weights.

Specifically, the main contributions of our work are:
\begin{enumerate}
    \item A non-parametric Bayesian approach for uncertainty estimation in DNN-based image registration models.
    \item Demonstration of uncertainty estimation that is highly correlated with the presence of out-of-distribution data compared to previously proposed DNN-based method for uncertainty estimation.
    \item Flexible approach that can be adjoined to most DNN training schemes, either supervised or unsupervised and architectures.
    \item Improved registration accuracy and smoothness as well as generalization capacity compared to baseline DNN method for diffeomorphic image registration. 
\end{enumerate}

The remainder of this paper is organized as follows: section \ref{sec:back} provides a general background on image registration. Then, we introduce the proposed NPBDREG method in section \ref{sec:NPBDREG}. Section \ref{sec:measures} describes the metric used to evaluate the registration performance and the measure that quantifies the voxel-wise uncertainty. Section \ref{sec:expresults} discusses the experiments performed on brain MRI images, and lastly, section \ref{sec:conc} concludes our work and discusses open questions and possible future studies.

\section{Background: Image Registration}\label{sec:back}
 Registration is the process of mapping a pair of images (e.g MR images acquired from different subjects, or from the same subject at different time-points) onto one coordinate system. Registration models can be loosely characterized by two categories of spatial transformations: \textit{global} and \textit{local}. The former is related to the rigid family, where an image is translated or rotated, or to the affine transformations, which perform shear mapping and scaling, in addition to rotation and translation while the latter accounts for non-linear dense transformations, or spatially varying (non-uniform) deformation models \cite{sotiras2013deformable}. Local deformations can be modeled by various geometric, physical, and interpolation models (e.g the cubic B-splines model \cite{rueckert1999nonrigid}). For comprehensive reviews we refer the reader to \cite{haber2004numerical,sotiras2013deformable}.
 
Classical approaches formulate the deformable registration task as an optimization problem. Let us denote the pair of fixed and moving images by $I_{f}$ and $I_M$, respectively. $\Phi$ is the deformation field, which accounts for mapping the grid of $I_M$ to the grid of $I_f$. Then, we estimate $\Phi$ by minimizing the following energy functional:
\begin{equation}
\hat{\Phi}=\argmin_\Phi {S(I_f,I_M\circ\Phi)+\lambda R(\Phi)} \label{eq:losseq}
\end{equation}
where $I_M\circ\Phi$ denotes the result of warping the moving image with $\Phi$. $S$ is a dissimilarity term, which quantifies the resemblance between the resulting image and the fixed input, and $R$ is a regularization term that encourages the deformation smoothness. The scalar $\lambda$ is a tuning-parameter that accounts for balancing the two terms, and it controls the smoothness of the resulting deformation. 

In DNN-based approaches, the registration task is translated to a prediction by a trained DNN:
\begin{equation}
\hat{\Phi} = f_{\theta}(I_{M},I_{F})
\end{equation}
where $I_{M},I_{F}$ are the input pair of images and $\theta$ are the weights of the DNN that are obtained after a training process which optimizes the following: 
\begin{equation}
\hat{\theta}= \argmin_\theta {S(I_F,I_M\circ f_\theta(I_{M},I_{F}))+\lambda R(f_\theta(I_{M},I_{F}))} \label{eq:lossdeep}
\end{equation}  
This kind of optimization provides the best point-estimate of the DNN weights, $\hat{\theta}$, that is the solution of Maximum Likelihood Estimation (MLE) rather than characterizing the entire posterior distribution of the DNN weights which is required for the assessment of the uncertainty involved in the predicted registration. Further, it may result in a potential over-fitting which may yield sub-optimal results at inference phase, especially for small-size datasets frequently present in the medical imaging domain.

In contrast, Bayesian DNN-based models have the potential to characterize the entire posterior distribution of the DNN weights and thus provide meaningful measures of the uncertainty involved in the prediction \cite{neal2012bayesian,cheng2019bayesian}:
\begin{equation}
    P(\theta|I_M,I_F) = \frac{P(I_M,I_F|\theta)P(\theta)}{\int_\Theta P(I_M,I_F|\theta')P(\theta')\:d\theta'} 
\end{equation}
Since direct integration of the posterior distribution is intractable, it is common to assume either a general or a domain-specific formulation of the prior $P(\theta)$ such as a Gaussian distribution with $\mu=0$ on the DNN weights and aimed to maximize the posterior estimation of the prediction:
\begin{equation}
\hat{\theta}= \argmax_\theta {P\left(\theta|I_M,I_F\right) \propto P(I_M,I_F|\theta)P(\theta)} \label{eq:basreg}
\end{equation} 
While this can provide a maximum-a-posterior (MAP) point estimation of the DNN output, it cannot characterize the entire distribution of the output, which is necessary to assess the uncertainty in the DNN prediction. 

\section{The NPBDREG Approach} \label{sec:NPBDREG}
To address this challenge, we introduce a non-parametric approach to characterize the actual posterior distribution of the DNN weights. Specifically, we treat the DNN weights as random variables and use an SGLD mechanism \cite{welling2011bayesian} to sample the posterior distribution of the DNN weights as follows. We incorporate a noise scheduler that injects a time-dependent Gaussian noise to the gradients of the loss during the training procedure. At every training iteration, we add Gaussian noise with \textit{adaptive} variance to the loss gradients. The weights are then updated in the next iteration according to the ``noisy'' gradients. This noise schedule can be performed with any stochastic optimization algorithm. In this work we focused on the formulation of the method for the \texttt{Adam} optimizer. We use Gaussian noise with a variance proportional to the learning rate of the \texttt{Adam} optimization algorithm, as it allows the adaption of the noise to the nature of loss curve. Moreover, previous research shows better performance using adaptive or time-dependent variance than constant variance \cite{neelakantan2015adding}. Our SGLD-based registration NPBDREG method is outlined in algorithm \ref{alg:1}. 
\begin{algorithm}[ht!] 
	\SetAlgoLined
	\SetKwFunction{Ftrain}{TrainNetwork}
	\SetKwFunction{Ftest}{FeedForward}
	\caption{NPBDREG Algorithm}\label{alg:1}
	\SetKwInput{KwInput}{Input}                
	\SetKwInput{KwOutput}{Output}
	\KwInput{Tuning parameters $\lambda$ and $\alpha$, and number of epochs $N$}
	\KwOutput{Estimated deformation $\hat{\Phi}$ and registered image $I_R$}
	\KwData{Dataset of pairs of fixed and moving images $I_M,I_F$}	
	\tcc{\underline{The \textit{offline} training}} 
	\SetKwProg{Fn}{Function}{:}{}
	\Fn{\Ftrain{${I_M,I_F}$,$\lambda$,$\alpha^t$,$t_b$,$N$}}
	{
		\For{$t<N$}
		{   Compute loss, $L^{t}$, for validation and training sets
			$\tilde{g}^{t}\leftarrow g^{t}+\textbf{N}^{t}$, where $\textbf{N}^{t}\sim\mathcal{N}(0,\frac{s^{t}}{\alpha^t})$
			and $s^{t}$ is the \textit{adaptive} step size. \\
			$\theta^{t+1}\leftarrow \text{\textsc{Adam\_Update}}(\theta^{t},\tilde{g}^{t})$ \\

		}
		\KwRet{$\left\{\theta^{t}\right\}_{t_{b}}^{N}$,$\left\{L^t\right\}_{t_{b}}^{N}$ } \\
	}
	\setcounter{AlgoLine}{0}
	\tcc{\underline{\textit{Online} Registration}} 
	\Fn{\Ftest{$I_M$,$I_F$,$\left\{\theta_{t}\right\} _{t_{b}}^{N}$,$\left\{w^t\right\}_{t_{b}}^{N}$}}
	{
		Compute a set of velocity fields $\left\{V^{t}\right\} _{t_{b}}^{N}$ by feed-forwarding $I_M$, $V^{t}=f_{\theta^t}(I_M,I_F)$ for $t\in \left[t_b,N\right]$ \\
		Estimate the weighted mean: 
		$\hat{V}=\frac{\sum_{t=t_b}^{N}w^{t}V^t}{\sum_{t=t_b}^{N} w^{t}}$\\
		$\Sigma_v \leftarrow$ The variance of $\hat{V}$ given $\left\{ V^{t}\right\} _{t_{b}}^{N}$ \\
		Calculate the difffeomorphic deformation: $\Phi\leftarrow\text{\textsc{Int\_Layer}}(I_{in},\Phi)$ \\
		Register image: $I_{R}\leftarrow\text{\textsc{Spatial\_Transform}}(I_{in},\Phi).$ \\
		
		\KwRet{$I_R,\Phi,\Sigma_v$}
	}
\end{algorithm}  
  
\subsection{Offline Training}\label{subsec:offlinetraining}
Let $L(I_M,I_F,f_\theta(I_M,I_F))$ denote the overall registration loss, described in \eqref{eq:lossdeep}, which is composed of both similarity and regularization terms. We denote the loss gradients by:
\begin{equation}
g^t\overset{\triangle}{=}\nabla_{\theta}L^t(I_{M},I_{F},f_{\theta}(I_{M},I_{F}))
\end{equation}
where $t$ is the training iteration (epoch). At each training iteration, Gaussian noise with time-varying variance is added to $g$:
\begin{equation}
\tilde{g}^{t}\leftarrow g^{t}+\textbf{N}^{t} 
\end{equation}
where $\textbf{N}^{t}\sim\mathcal{N}(0,\frac{s^{t}}{\alpha^{t}})$, $s^{t}$ is the \texttt{Adam} step size, and $\alpha^{t}$ is a user-selected parameter that controls the noise variance (can be time-decaying or a constant). This is especially important in the first learning stages, which involve a large step size. 
The network parameters are then updated according to the \texttt{Adam} update rule:
\begin{equation} 
\theta^{t+1}\leftarrow \theta^{t} - s^{t}\hat{m}^{t}
\end{equation}
where $s^{t}=\frac{\eta}{\sqrt{\hat{v}^{t}+\epsilon}}$, $\hat{m}^{t}$ and $\hat{v}^{t}$ are the bias-corrected versions of the decaying averages of the past gradients and the second moment (squared gradients), respectively:
\begin{align}
\hat{m}^{t} = \nicefrac{m^{t}}{1-\beta_{1}^{t}} \nonumber \\
\hat{v}^{t} = \nicefrac{v^{t}}{1-\beta_{2}^{t}}
\end{align}
where $m^{t}=m^{t-1}+(1-\beta_{1})\tilde{g}^{t}$ and $v^{t}=v^{t-1}+(1-\beta_{2})(\tilde{g}^{t})^2$. $\beta_{1}$, $\beta_{2}$ are decay rates and $\eta$ is a fixed constant. 
Lastly, we save the weights of the network that were obtained in iterations $t\in \left[t_b,N\right]$, where $t_b$ is a pre-determined  parameter of the NPBDREG method and $N$ is the overall number of iterations. It is essential to select a $t_b$ larger than the cut-off point of the \textit{burn-in} phase. One should sample weights obtained in the last $t_b,..,N$ iterations, where the loss curve has converged. Under some feasible constraints on the step size, the sampled weights converge to the posterior distribution \cite{welling2011bayesian} (See \ref{app:1} for further discussion). 
Thus, outputs of our network after the \textit{burn-in} phase, trained with the {\it adaptive} SGLD approach, can be considered a sampling from the true posterior distribution.

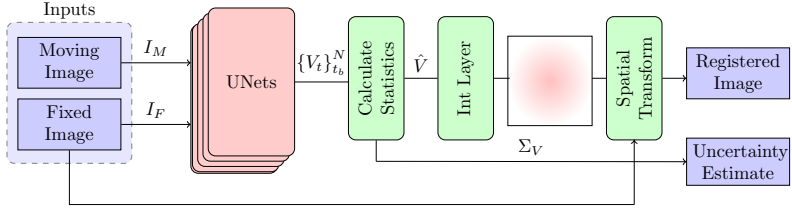
\begin{figure*}[t]
\scalebox{0.7}{	
\centering{
		\pgfdeclarelayer{background}
		\pgfdeclarelayer{foreground}
		\pgfsetlayers{background,main,foreground}
		
		\tikzstyle{sensor}=[draw, fill=blue!20, text width=4.9em, 
		text centered, minimum height=2.5em]
		\tikzstyle{ann} = [above, text width=4em]
		\tikzstyle{naveqs} = [sensor, text width=4em, fill=red!20, 
		minimum height=8em, rounded corners]
		\tikzstyle{naveqs1} = [sensor, text width=6em, fill=green!20, 
		minimum height=3em, rounded corners]
		
		\def\blockdist{2.3}
		\def\edgedist{3}
		
		\begin{tikzpicture}
		\node (naveq) at (0,0) [naveqs] {UNet};
		\node (naveq1) at (0.03,0.03) [naveqs] {UNets};
		\node (naveq2) at (0.13,0.13) [naveqs] {UNets};
		\node (naveq3) at (0.23,0.23) [naveqs] {UNets};
		\node (naveq4) at (0.33,0.33) [naveqs] {UNets};
		\node (naveq41) at (2.7,0.4) [naveqs1,rotate=90] {Calculate Statistics};
		\node (naveq5) at (4.4,0.4) [naveqs1,rotate=90] {Int Layer};
		\node (naveq6) at (7.6,0.4) [naveqs1,rotate=90] {Spatial Transform};
		
		\path (naveq.140)+(-\blockdist,0) node (gyros) [sensor] {Moving Image};
		\path (naveq.-150)+(-\blockdist,0) node (accel) [sensor] {Fixed Image};
		\path (naveq.140)+(10.4,-0.2) node (reg) [sensor] {Registered Image};
		\path (naveq.140)+(10.4,-1.88) node (sigma) [sensor] {Uncertainty Estimate};
		
		\path [draw, ->] (gyros) -- node [above] {$I_M$} 
		(naveq.west |- gyros) ;
		
		\path [draw, ->] (accel) -- node [above] {$I_F$} 
		(naveq.west |- accel);
		
		\node (IMU) [above of=gyros] {Inputs};
		\path (naveq.south west)+(-0.6,-0.4) node (INS) {};
		
		\draw[-] (naveq4) -- node [above](\edgedist,0) {{$\left\{ V_{t}\right\} _{t_{b}}^{N}$}} ( naveq41.north |- naveq4);
		\draw [-] (naveq41) -- node [above](\edgedist,0) {$\hat{V}$} ( naveq5.north |- naveq41);

		\shadedraw[inner color=pink,outer color=white, draw=black] (5.2,-0.5) rectangle +(1.6,1.7);
		\draw (naveq5.south) -- node[above] {} ++(0.3,0);
		\draw[-] (6.8,0.4) |- (naveq6.north);
        \draw[-] (naveq41.west) -- node[below] {} ++(0,-0.4);
        \draw[->] (naveq41.west)++(0,-0.4) -- node[above] {$\Sigma_{V}$} (sigma.west);

		\draw[-] (accel.south)++(0,0) |- (7.6,-2);
		\draw[->] (7.6,-2) |- (naveq6.west);
		\draw [->] (naveq6) -- node [above](\edgedist,0) {} (reg.west |- naveq6);
		
		\begin{pgfonlayer}{background}
		\path (gyros.west |- naveq.north)+(-0.2,0.05) node (a) {};
		\path (INS.south -| naveq.east)+(+0.25,-0.25) node (b) {};
		\path (gyros.north -| gyros.east)+(+0.2,-2.4) node (b) {};
		\path[fill=blue!10,rounded corners, draw=black!50, dashed]
		(a) rectangle (b);
		\end{pgfonlayer}
		
		\end{tikzpicture}
        }   }	
	\caption{Block diagram of the proposed NPBDREG system. The result of the training procedure is a set of backbone models characterizing the posterior distribution of the DNN weights. At inference, we sample from the posterior distribution of the predicted velocity fields by passing the fixed and moving images through each one of the backbone models.} After having a set of velocity fields, $\left\{ V_{t}\right\} _{t_{b}}^{N}$, we calculate the averaged fields and the voxel-wise variance, $\hat{V}$ and $\Sigma_{V}$, respectively. The average is used as the most probable registration and the $\Sigma_{V}$ is utilized for uncertainty assessment. 
	\label{fig:bddiagram}
\end{figure*}

\subsection{The Registration System}
 Fig.~\ref{fig:bddiagram} illustrates the operation of NPBDREG on a pair of fixed and moving images. During the training of our backbone Probabilistic \texttt{VoxelMorph} model \cite{dalca2018varreg} as described in Sec.~\ref{subsec:offlinetraining}, we sample a set of weights that were obtained in the last $t\in \left[t_b,N\right]$ iterations.
   At inference, we feed-forward pairs of fixed and moving images ${I_M,I_F}$ through the DNNs to obtain a a set of velocity fields $\left\{ V^{t}=f_{\theta^t}(I_M,I_F)\right\}_{t_{b}}^{N}$. 
Next, we estimate the averaged posterior velocity field:
\begin{equation}
\hat{V}=\frac{\sum_{t=t_b}^{N} w^{t}V^t}{\sum_{t=t_b}^{N} w^{t}}
\end{equation} 
where $w_{t}$ is a predetermined weight parameter. To give a larger weight to networks with better performance, $w^{t}$ is proportional to the total loss calculated on the validation set, at epoch $t$. This is due to the fact that the loss is negative where local-cross-correlation (LCC) was used as a dissimilarity loss. In addition, we quantify the voxel-wise uncertainty maps of the registration by means of empirical variance of the velocity fields, $\Sigma_V$. A numerical integration, using scaling and squaring (briefly outlined in subsection \ref{subsec:intlayer}), operates on the averaged velocity field, $\hat{V}$, to yield a diffeomorphic deformation, $\Phi$.
Lastly, we register the moving image by resampling its coordinate system with the spatial transform $\Phi$. The function \textsc{Spatial\_Transform} performs spatial warping as in \cite{dalca2018varreg}. For each voxel $p$, a sub-voxel location $\Phi(p)$ is calculated. Then, the values are linearly interpolated to obtain an integer. 

\paragraph{Uncertainty Estimation}
Uncertainty of DNN models can be loosely classified into two types: Aleatoric and Epistemic \cite{kendall2017uncertainties}. The former is caused by noise in the data (for example, noise added to the input images), whereas, the latter is induced by an uncertainty in the parameters of the model. In DNN frameworks, Bayesian methods aim to quantify the Epistemic uncertainty by characterizing the posterior predictive distribution. As such, our NPBDREG system provides a voxel-wise Epistemic uncertainty map of the registration by sampling from the posterior of our network. Particularly, we sample a set of velocity fields from DNNs with weights obtained in iterations $t_b,..,N$, $\left\{ V_{t}\right\} _{t_{b}}^{N}$. We then calculate the voxel-wise diagonal covariance (variance), $\Sigma_{V}$. Finally, the uncertainty of the velocity field can be calculated by:
\begin{equation}
    H_{V} = \frac{1}{2}\log(2\pi\Sigma_{V}) \label{eq:Hcalc}
\end{equation}
By the same manner, one may quantitatively assess the uncertainty of the registration field where $\Sigma_{V}$ in \eqref{eq:Hcalc} is replaced by the diagonal covariance calculated over the set of Diffeomorphic Deformations obtained by \textit{scaling of squaring} each $\left\{ V_{t}\right\} _{t_{b}}^{N}$.

\begin{figure}
    \centering
    \includegraphics[width=6cm]{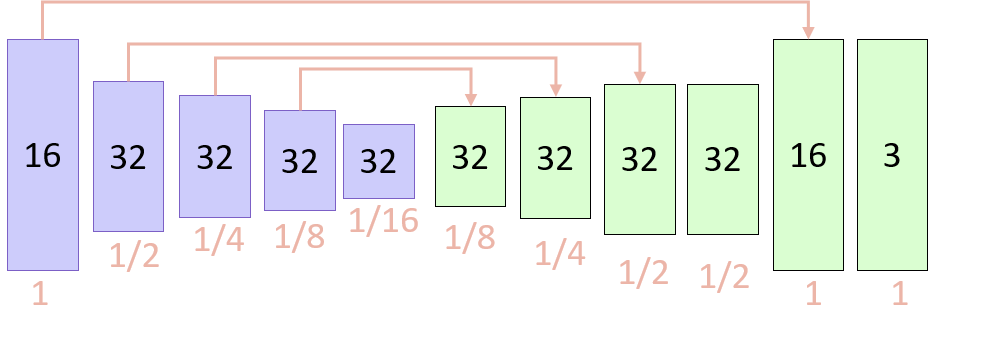}
    \caption{The backbone DNN-based registration model of our NPBDREG system. Arrows represent skip connections between the corresponding layers. CNN layers of the encoder and decoder are highlighted in purple and green blocks, respectively. Downsample and upsample ratios are written underneath each block.}
    \label{fig:netarch}
\end{figure}

\paragraph{Backbone architecture and Training}
The backbone of our registration system is based on the UNet architecture, similar to the Probabilistic \texttt{VoxelMorph} model \cite{dalca2018varreg}. It  comprise of an encoder and decoder with skip connections, as highlighted in Fig.~\ref{fig:netarch}. Both encoder and decoder parts consist of CNN layers with kernel size $3\times3\times3$, followed by Leaky ReLU activation functions. The encoder has $5$ CNN layers each with $\left\{16,32,32,32,32\right\}$ channels (blocks in purple, Fig.~\ref{fig:netarch}), whereas, the decoder consists of $5$ layers with the following number of channels: $\left\{32,32,32,32,16\right\}$ (blocks in green, Fig.~\ref{fig:netarch}). Afterwards, a 3-channel convolutional layer is applied to the UNet output to obtain the 3D deformation field.
The training process of the model involves the optimization of the energy-functional described in \eqref{eq:lossdeep}. We use LCC for our MRI images experiments to characterize the dissimilarity between the fixed image $I_{F}$ and the registered image obtained after mapping, $I_{M}\circ\hat{\Phi}$. The $\text{L}_{2}$ norm over the deformation field gradients is used as a regularization term to encourage the deformation field smoothness. The UNet model then learns the optimal weights by minimizing the loss function composed of the two aforementioned terms.   
\subsection{Integration method} \label{subsec:intlayer}
Diffeomorphic deformation is a differentiable and invertible transformation, $\Phi:\mathbb{R}^3\rightarrow \mathbb{R}^3$, which maps the coordinates from the moving image to the fixed image. Similarly to \cite{dalca2018varreg}, we consider a stationary velocity field representation, where 
the deformation field is defined through the following ODE:
\begin{equation}
    \begin{cases}
    \begin{array}{c}
    \frac{\partial\Phi^{(t)}}{\partial t}=V(\Phi^{(t)})\\
    \Phi^{(0)}=\boldsymbol{I}
    \end{array}\end{cases} \label{eq:ode}
\end{equation}
where $V$ is the deformation field and $I$ is the identity transformation. Then, We obtain $\Phi$ by integrating the stationary velocity field, $v$, over $t=\left[0,1\right]$. The solution of the ODE \eqref{eq:ode} is characterized by an exponential of $V$, that is  $\Phi^{(1)}=exp(V)$. In our implementation we adopt the \textit{Scaling and Squaring} method to compute the matrix exponential efficiently \cite{arsigny2006log}. The \textit{scaling and squaring} scheme can be outlined in the following steps:
\begin{enumerate}
    \item  \textbf{Scaling step:} $V(x) \leftarrow \frac{V(x)}{2^T}$ \\
    where $\frac{V(x)}{2^T}$ is sufficiently small (desired accuracy according to predetermined $T$, we refer to $T$ as the integration steps) 
    \item  \textbf{Exponentiation step:} $\Phi^{\frac{1}{2^T}}\leftarrow V(x)$ (first-order explicit scheme)
    \item \textbf{Squaring step:} Perform $T$ recursive squarings of $\Phi^{(\frac{1}{2^T})}$\\
    loop over $t=\{T,...,1\}$: $\Phi^{\frac{1}{2^{t-1}}} \leftarrow \Phi^{\frac{1}{2^t}} \circ \Phi^{\frac{1}{2^t}}$
\end{enumerate}
The intuitive interpretation of this method is that the deformation at time $T=1$ is given as a result of
$2^T$ times the composition of the small deformations obtained at time $\nicefrac{1}{2^T}$. For further details about this method refer to \cite{arsigny2006log}.

\section{Experimental Methodology}\label{sec:measures}
\subsection{Database and Preprocessing}
T1-weighted MRI images of the brain for $80$ subjects were acquired from four different database sources: MGH10, CMUC12, ISBR18 and LPBA40.\footnote{used in \url{https://continuousregistration.grand-challenge.org/data/} and in \cite{klein2009evaluation} (with a detailed description about each database.)}
MGH10 consists of the brain MRI images of $10$ subjects, with atlases of segmentations divided into $74$ manually labeled regions. The images were inhomogeneity-corrected and affine-registered to the MNI152 template. CMUC12 includes MRIs of $12$ subjects with atlases of $128$ labeled regions. The images were rotated into cardinal orientation. ISBR18 consists of T1-weighted MRI image data of $18$ subjects and their segmentations were divided into $84$ individual anatomical structures. Images were processed by autoseg bias field correction. LPBA40\footnote{are available at \url{http://www.loni.ucla.edu/Atlases/LPBA40}} has whole-head MRI images of $40$ subjects with manual delineations of $56$ structures \cite{shattuck2008construction}. Images were preprocessed according to existing protocols to produce skull-stripped brain volumes.
All images were multiplied by their provided brain masks, zero-padded to a size of $256\times 256\times 256$, resampled to a uniform grid and size ($1\times1\times1$), then cropped to the central $160\times 192 \times 224$, and normalized to the $\left[0,1\right]$ gray-scale domain. Further, affine alignment was performed using \texttt{dipy} library, which maximizes the mutual information (MI) via optimization strategy similar to ANTS \cite{avants2011reproducible}.
We split each dataset into: training, evaluation, and test sets with the following ratios: $56\%$, $30\%$ and $14\%$, respectively.
Then, we constructed  $292$, $20$ and $78$ pairs from each dataset for training, evaluation, and testing, respectively.
\subsection {Implementation Details and hyperparameters selection}
We implemented our NPBDREG method for brain MRI images using Keras with a tensorflow backend \cite{abadi2016tensorflow}.\footnote{source code of NPBDREG, which includes the training algorithm and preprocessing is available at: \url{https://github.com/samahkh/NPBDREG_prj}. } 
The NPBDREG network was trained with noise injections for $4,000$ iterations. We used an \texttt{Adam} optimizer \cite{kingma2014adam} with  a learning rate set to $2^{-4}$ and added a $L_{2}$ regularization of network weights and biases to our loss function to avoid overfitting. We empirically set the $L_{2}$ regularization weight to $10^{-7}$ and the total integration steps to, $T=6$. 
We conducted an exhaustive search to determine the best value for the smoothness parameter, $\lambda$ which provides  best performance in terms of both registration accuracy and smoothness of the deformation fields. 
We assessed the registration accuracy by means of the Dice score \cite{dice1945measures} between the reference segmentation on the fixed image and the propagated (registered) segmentation from the moving image. We quantified the registration smoothness by calculating the percentage of folds, i.e. voxels with negative determinant of the Jacobian, $|J_\Phi|<0$. 
Additionally, we trained the NPBDREG with two configurations: (1) a fixed $\alpha^t$, the std of the noise injected to the gradients was set to $s^t=\nicefrac{lr}{50}$. (2) a time-decaying $\alpha^t$, the std of the noise injected to the gradients was set to $s^t=\nicefrac{lr}{(1+t)^{0.55}}$. In our experiments, we selected $t_b=3392$ in algorithm~\ref{alg:1} and saved the weights obtained in the last $8$ iterations.
 At inference time, our model registers a new pair, $I_F$ and $I_M$, onto one coordinate system by resampling with the mean transformation $\Phi$, yielded by averaging deformations obtained in the last $8$ iterations. 
 We used the probabilistic \texttt{VoxelMorph} (PrVXM) model \cite{dalca2018varreg} as a baseline for the purpose of performance comparison in terms of registration accuracy and uncertainty quantification. We trained PrVXM with the same settings and loss mentioned in \cite{dalca2018varreg}. We used an \texttt{Adam} optimizer without gradient noise injections, as well as a UNet architecture identical to NPBDREG and the same loss functions. Both PrVXM and NPBDREG models have the same amount of trainable parameters ($265,237$), as they are composed from the UNet-based architecture. Additionally, in terms of computational complexity, both NPBDREG and PrVXM methods have the same straining time and runtime for registration prediction. However, to enable uncertainty quantification by sampling from the posterior of both models, one performs $8$ forward passes to the registration system, which, in turn, increases the overall runtime by factor of $8$. The uncertainty estimation runtime, therefore, is also about the same for both approaches.

\subsection{Evaluation of Registration Accuracy and smoothness}
We evaluated the performance of our Bayesian unsupervised registration system by means of registration accuracy and deformation smoothness. We used the same measures and in our hyper-parameter selection experiments, i.e. the Dice score as a measure of registration accuracy and the percentage of folds in the predicted deformation field as a measure of smoothness. 
We selected 6 labels of anatomical structures, provided by the atlases to evaluate the registration performance by means of Dice score. Firstly, initial Dice score after affine alignment was calculated for all labels in the atlases. Then, only $6$ labels with the largest volume size and a Dice score bigger than $0.5$ were selected and used in our experiments. The anatomical labels used for Dice calculations are presented in Table~\ref{tab:anatomicallabels} (\ref{appendix:details}).

\subsection{Improved Generalization Capability}
We analyzed the improvement in DNN-based registration generalization capabilities achieved by our NPBDREG approach by means of robustness to the presence of noise. We corrupted the input images with two types of noise: Gaussian noise with various std ($\sigma$) and mixed structures, which generated a linear combination of the test example ($i$) and another example sampled randomly from the test set ($j$): $\alpha I{j}+(1-\alpha)I_{i}$ \cite{freiman2019unsupervised}. 
We used the Dice score and the same labels as in the previous experiment to determine the added-value of our NPBDREG approach in increasing generalization capabilities of the registration system. 

\subsection{Uncertainty Assessment} \label{sebsec:uncasses}
 We assessed the capability of our NPBDREG approach to characterize uncertainty related to out-of-distribution data. We randomly selected  a group of $8$ images from the test set. We simulated out of distribution data by corrupting the images with Gaussian noise with varying $\sigma_n$. 
We computed the empirical voxel-wise diagonal covariance (variance), $\Sigma_V$ as follows: We sampled $8$ deformations, one from each of the networks with weights obtained in the last $8$ iterations. Then, we computed the uncertainty in the velocity field as mentioned in \eqref{eq:Hcalc}.
Finally we assessed the correlation between the computed uncertainty and the noise levels ($\sigma_n$).

\begin{figure*}
	\begin{minipage}[b]{0.9\linewidth}
		\centering
			\includegraphics[width=9cm]{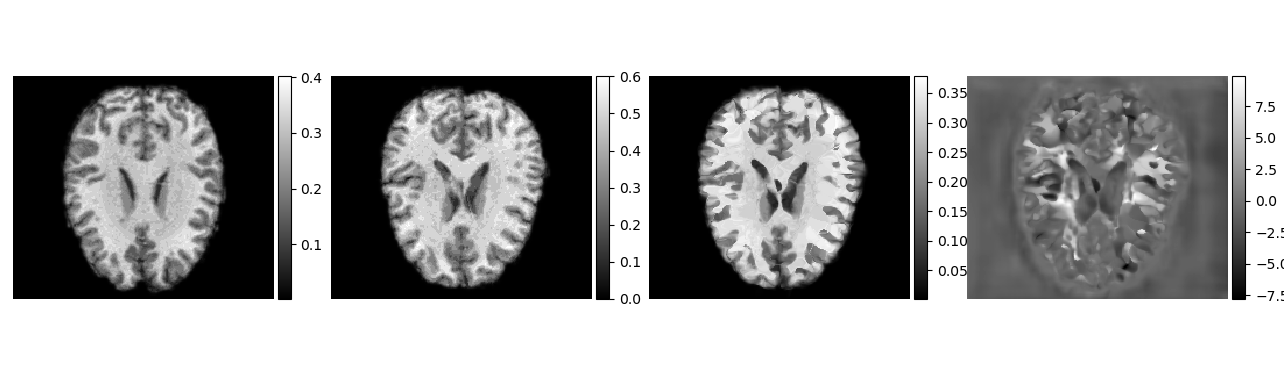}
	\end{minipage} \\
	\begin{minipage}[b]{0.9\linewidth}
		\centering
			\includegraphics[width=9cm]{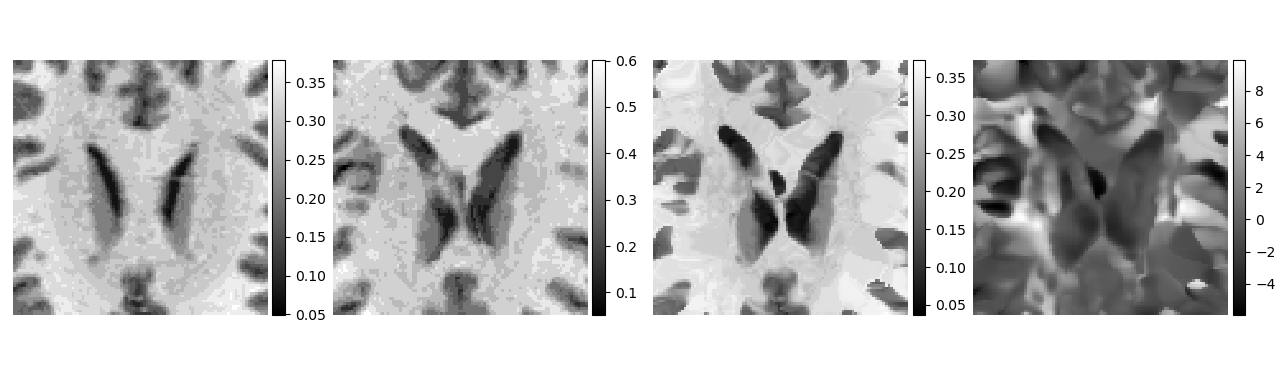}
	\end{minipage}
	\\
	\begin{minipage}[b]{0.9\linewidth}
		\centering
	
			\includegraphics[width=9cm]{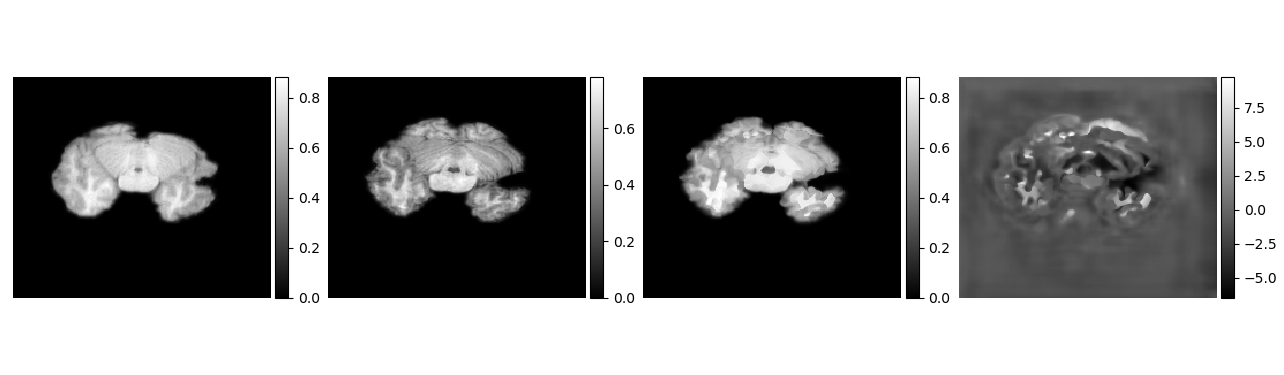}
	\end{minipage} \\
	\begin{minipage}[b]{0.9\linewidth}
		\centering
			\includegraphics[width=9cm]{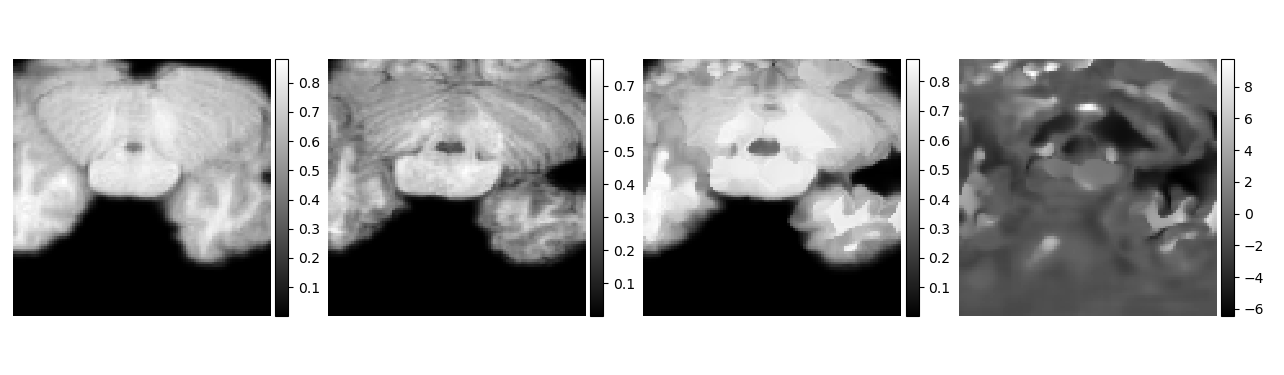}
	\end{minipage} \\
		\begin{minipage}[b]{0.9\linewidth}
		\centering
	
			\includegraphics[width=9cm]{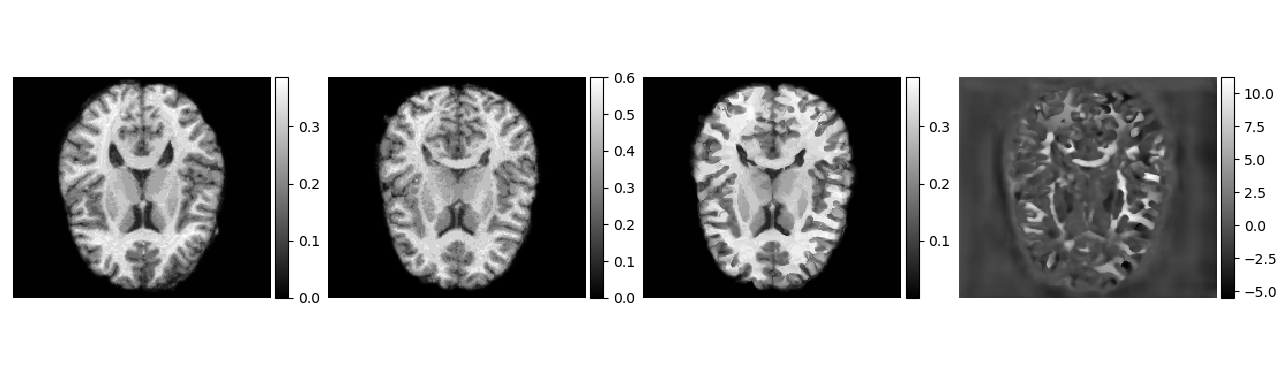}
	\end{minipage} \\
	\begin{minipage}[b]{0.9\linewidth}
		\centering
			\includegraphics[width=9cm]{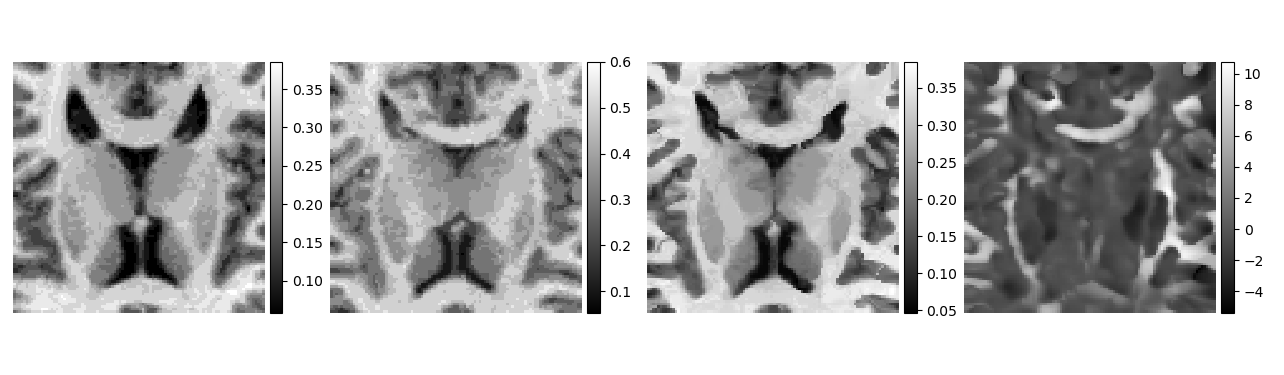}
	\end{minipage}
	\caption{Registration results. Rows $1$, $3$ and $5$: examples of MRI slices of input moving image, fixed image, and resulting warped image and the corresponding deformation field (from left to right). The observed images belong to MGH10 and LPBA40 database sources, respectively. Rows $2$, $4$ and $6$: closeups of the center regions of the moving, fixed and registered images, and the registration field, respectively.   }\label{fig:RegMRIexample}	
\end{figure*} 
\begin{table}[ht]
\centering
\scalebox{0.9}{
\begin{tabular}{c|c|c|c|c}
\hline
          & \multicolumn{2}{c|}{\textbf{$|J_\Phi|<0\quad\%$}} & \multicolumn{2}{c}{Dice} \\ \hline \hline
          \multicolumn{1}{l|}{}        & mean             & std               & mean             & std             \\ \hline
\textbf{$\lambda=0.005$} & 0.028            & 0.0041            & 0.725          & 0.0541          \\ \hline
\textbf{$\lambda=0.01$}  & 0.021            & 0.0034            & 0.735           & 0.0541          \\ \hline
\textbf{$\lambda=0.1$}   & \textbf{0.014}            & 0.0023            & \textbf{0.739 }          & 0.0537          \\ \hline
PrVXM                & 0.017            & 0.0043            & 0.688           & 0.0657          \\ \hline
\end{tabular}}
\caption{Percentages of folds and Dice score for different values of $\lambda$. The mean and std values of percentage of folds and Dice score, calculated over the whole test set, for the deformation fields obtained by NPBDREG with three values of $\lambda$, $\lambda=\{0.005,0.01,0.1\}$ and the baseline PrVXM. The number of folds i.e. voxels with negative determinant of the Jacobian, $|J_\Phi|<0$, is computed. Then, the percentage of folds for each volume is calculated by dividing the total number of folds by the overall number of voxels. The best Dice score and minimal number of folds are obtained by the model with $\lambda=0.1$. }\label{tab:tuninglambda}
\end{table}
\begin{figure*}[ht]
	\begin{minipage}[b]{0.32\linewidth}
		\centering
		\subfloat[\label{fig3:a}\footnotesize $|J|$, NPBDREG, $\lambda=0.1$]{
			\includegraphics[width=4.5cm]{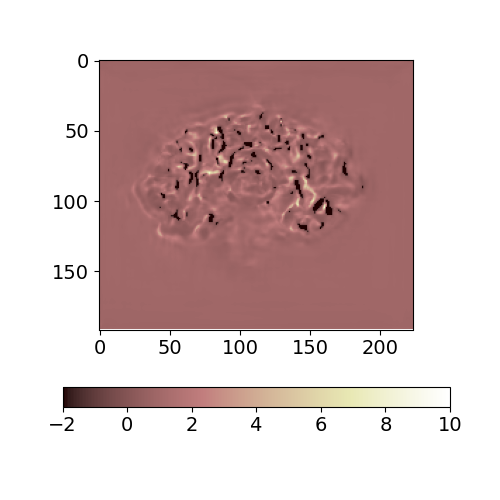}}
	\end{minipage}
	\begin{minipage}[b]{0.32\linewidth}
		\centering
		\subfloat[\label{fig3:b}\footnotesize $|J|$, NPBDREG $\lambda=0.01$]{
			\includegraphics[width=4.5cm]{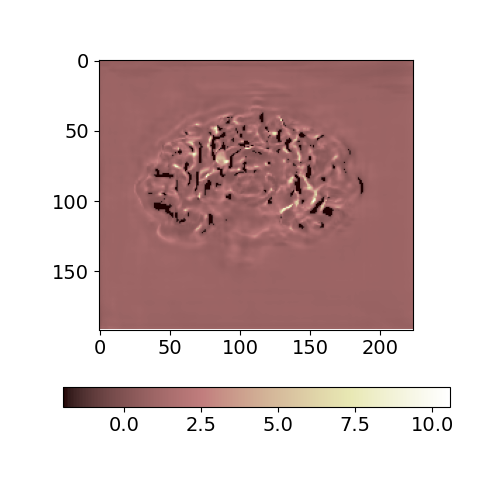}}
	\end{minipage}	
	\begin{minipage}[b]{0.32\linewidth}
		\centering
		\subfloat[\label{fig3:c}\footnotesize $|J|$, PrVXM]{
			\includegraphics[width=4.5cm]{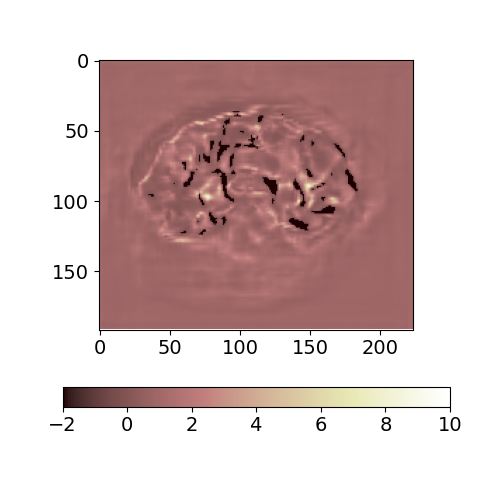}}
	\end{minipage}
	\caption{Hot maps of the Jaccobian determinant of the deformation field, $|J|$, obtained by NPBDREG with $\lambda=0.1$ \protect\subref{fig3:a}, NPBDREG with $\lambda=0.01$ \protect\subref{fig3:b} and PrVXM \protect\subref{fig3:b}. The percentage of folds for each example of the aforementioned deformations are: $0.013$, $0.019$ and $0.015$, respectively. The percentage of folds for each volume is calculated by dividing the total number of folds by the overall number of voxels.}\label{fig:DetDefs}	
\end{figure*} 

\section{Experimental Results}\label{sec:expresults}
\subsection{Hyperparameter selection}
The top 3 rows of table~\ref{tab:tuninglambda} presents the registration results of the NPBDREG model for 3 different values of $\lambda$ by means of registration accuracy and deformation smoothness averaged over the different anatomical labels and datasets. 
Fig.~\ref{fig:DetDefs} presents the Jaccobian determinant for the different values of $\lambda$. 
We selected the model $\lambda=0.1$ which optimized both the number of folds and Dice score. As one may expect, higher values $\lambda$ lead to better smoothness. However, the selection of $\lambda>1$ didn't boost the performance in terms of Dice score. Table~\ref{tab:implementdet} (\ref{appendix:details}) summarizes  the hyperparameters that were used in our experiments.

\subsection{Evaluation of Registration Accuracy and smoothness}
Fig.~\ref{fig:RegMRIexample} presents several representative registration results of our NPBDREG system.
The bottom 2  rows of table~\ref{tab:tuninglambda} presents the registration accuracy and smoothness in comparison to the baseline PrVXM \cite{dalca2018varreg} over the entire test set. Our NPBDREG ($\lambda=0.1$) improved overall registration accuracy by $\sim7.5\%$ ($0.739$ vs. $0.688$, $p \ll 0.01$, table~\ref{tab:pvalues}, first column) and registration smoothness by $\sim18\%$ ($0.014$ vs. $0.017$, $p\ll 1e^{-4}$).

Table~\ref{tab:dicedata} presents the averaged values of Dice metrics, over the anatomical labels, for each one of the datasets separately. Both NPBDREG with configurations (1) and (2) achieved improvements over PrVXM in terms of Dice metric. In addition, the nature of $\alpha^t$, which tunes the std of the gradients' noise, doesn't have a significant effect on the registration accuracy, i.e. both fixed and time-decaying $\alpha^t$ yielded approximately similar performance ($p=0.093$). Thus, in the following analysis and experiments we consider only NPBDREG (1), where $\alpha^t$ is fixed. 
 \begin{table*}[t!]
\centering
\scalebox{0.8}{
\begin{tabular}{c|l|l|l|l|l|l|l|l}
\hline
\textbf{Method}      &  \multicolumn{2}{c|}{\textbf{MGH10}} & \multicolumn{2}{c|}{\textbf{CUMC12}} & \multicolumn{2}{c|}{\textbf{ISBR18}} & \multicolumn{2}{c}{\textbf{LPBA40}} \\ \hline\hline
\textbf{NPBDREG (1)} & \multicolumn{1}{c|}{\textbf{0.798}} & 0.0101 & \textbf{0.7302}            & 0.0407          & 0.626           & 0.0068            & \textbf{0.742}            & 0.0506           \\ \hline
\textbf{NPBDREG (2)} & 0.790                      & 0.0089 & 0.721            & 0.0484            & \textbf{0.625 }           & 0.0154            & 0.725            & 0.0512            \\ \hline
\textbf{PrVXM}       & 0.797                     & 0.0107 & 0.629            & 0.0628            & 0.564            & 0.0392            & 0.694            & 0.0562            \\ \hline
\end{tabular}}
\caption{Registration accuracy evaluation results. The mean and std of Dice metric, calculated over labels and test examples for each database (columns), are presented for the three different models (from top to bottom: NPBDREG with configurations (1) and (2) and the baseline PrVXM, respectively.}\label{tab:dicedata}
\end{table*}
 \subsection{Improved Generalization Capability}
 Table~\ref{tab:dicenoise} summarizes the quantitative results obtained by NPBDREG with the two configurations and PrVXM, for different noise levels $\sigma$ and $\alpha$. NPBDREG  shows a significant improvement over PrVXM in noisy scenarios ($p \ll 0.01$ for most of the noise levels, see Table~\ref{tab:pvalues}). One may note that the accuracy of registration obtained by our method and PrVXM are similar for the MGH10 database. In our evaluation experiments, only a few examples from MGH10 were selected in the test set, since this source includes only $10$ subjects. 
\begin{table*}[t!]
\centering
\scalebox{0.75}{
\begin{tabular}{c|l|l|l|l|l|l|l|l|l|l}
\hline 
 \textbf{} & \multicolumn{6}{c|}{\textbf{Gaussian}} & \multicolumn{4}{c}{\textbf{Mixed Structure}}   \\ 
\hline
\textbf{Method}      & \multicolumn{2}{c|}{\textbf{$\sigma=0$}} & \multicolumn{2}{c|}{\textbf{$\sigma=0.4$}} & \multicolumn{2}{c|}{\textbf{$\sigma=0.6$}} & \multicolumn{2}{c|}{\textbf{$\alpha=0.2$}} & \multicolumn{2}{c}{\textbf{$\alpha=0.4$}} \\ \hline \hline
\textbf{NPBDREG (1)} & \textbf{0.739}             & 0.0537            & \textbf{0.702}              & 0.0568             & \textbf{0.682}              & 0.0566             & \textbf{0.729}              & 0.0563             & \textbf{0.711}              & 0.0605             \\ \hline
\textbf{NPBDREG (2)} & 0.723             & 0.0541            & 0.698              & 0.0548             & 0.681              & 0.0555             & 0.716              & 0.0558             & 0.7005             & 0.0602             \\ \hline
\textbf{PrVXM}       & 0.688             & 0.0657            & 0.687              & 0.0619             & 0.679              & 0.0623             & 0.686              & 0.0640             & 0.674              & 0.0659             \\ \hline
\end{tabular}}
\caption{Registration robustness to noise evaluation results. The added noise levels are denoted by $\sigma$ and $\alpha$, respectively. The mean and std of the Dice metric, calculated over the whole test set and over labels, are presented for the three different models (from top to bottom: NPBDREG with configurations (1) and (2) and the baseline PrVXM, respectively. }\label{tab:dicenoise}
\end{table*}
\begin{figure*}[t!]
	\begin{minipage}[b]{0.32\linewidth}
		\centering
		\subfloat[\label{fig4:a}\footnotesize $\hat{V}_{x}$]{
			\includegraphics[width=4.5cm]{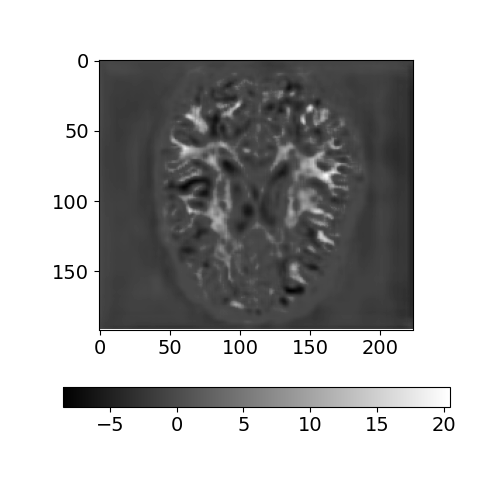}}
	\end{minipage}
	\begin{minipage}[b]{0.32\linewidth}
		\centering
		\subfloat[\label{fig4:b}\footnotesize $\Phi_{x}$]{
			\includegraphics[width=4.5cm]{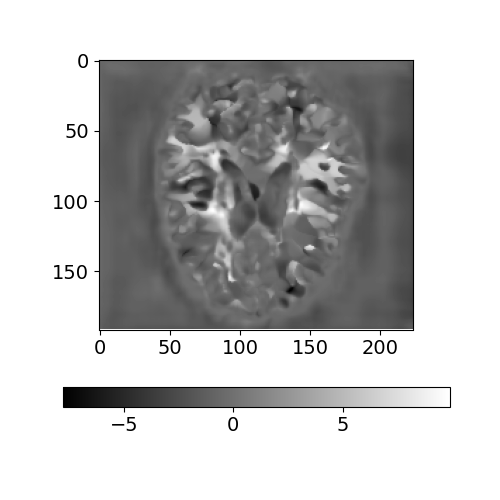}}
	\end{minipage}	
	\begin{minipage}[b]{0.32\linewidth}
		\centering
		\subfloat[\label{fig4:c}\footnotesize $H_{x}$]{
			\includegraphics[width=4.5cm]{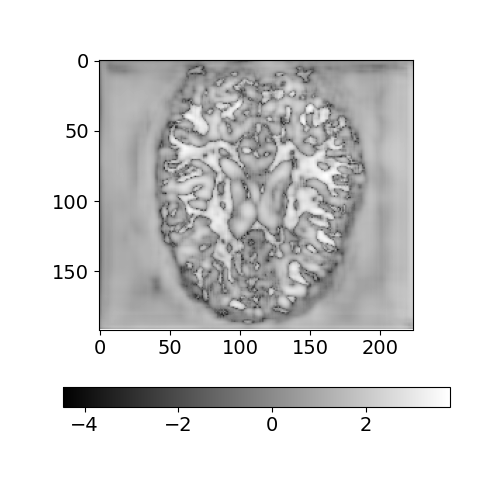}}
	\end{minipage}
	\caption{Uncertainty assessment. \protect\subref{fig4:a} and \protect\subref{fig4:b} The averaged velocity field and the corresponding deformation in x direction, respectively. \protect\subref{fig4:c} uncertainty maps of the deformation in x direction, $H_{x}$.}\label{fig:uncexample}	
\end{figure*}
\subsection{Uncertainty Assessment}
\begin{figure}[t!]
	\begin{minipage}[b]{0.45\linewidth}
		\centering
		\subfloat[\label{fig5:a}\footnotesize NPBDREG ]{
			\includegraphics[height=4.1cm]{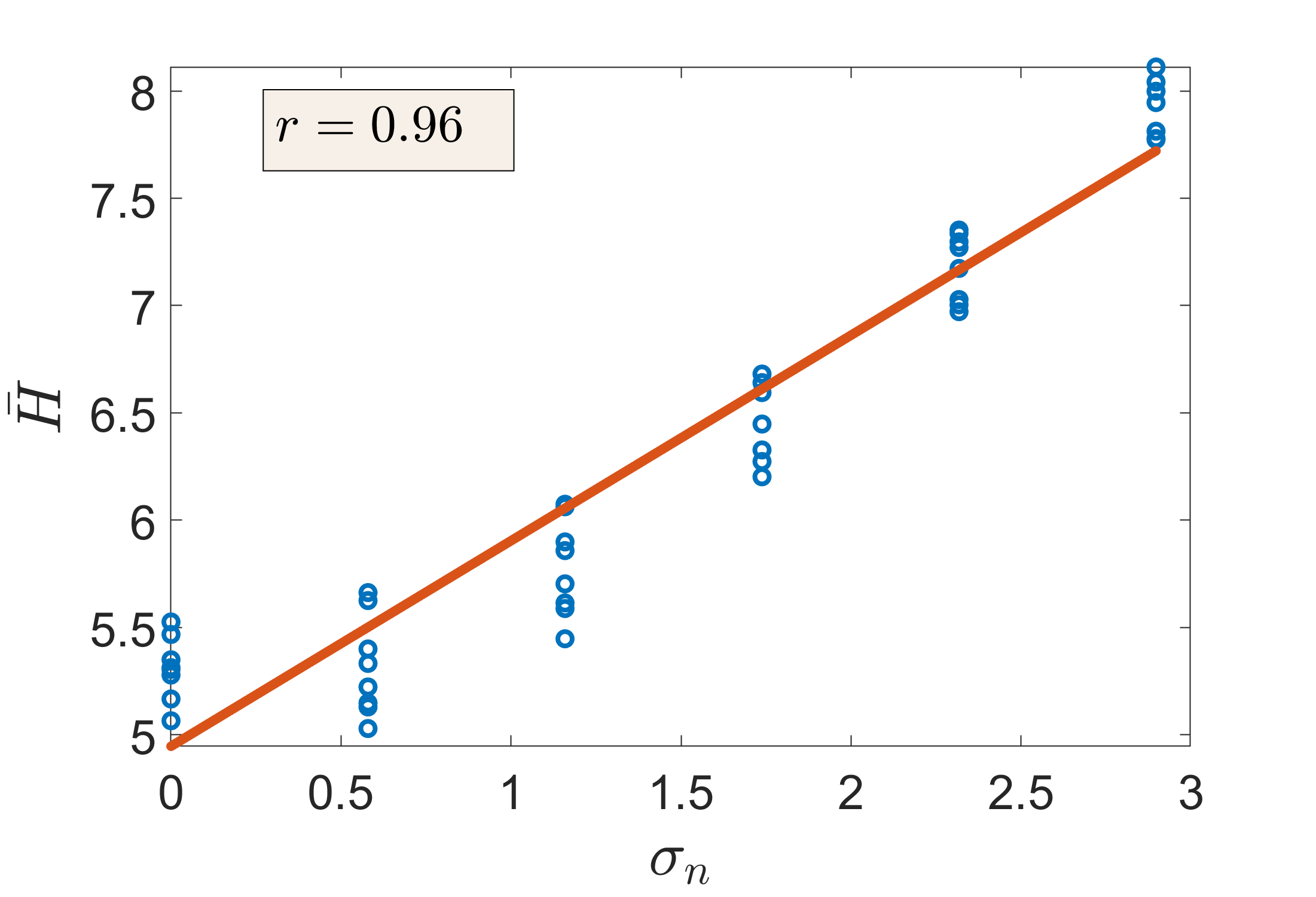}}
	\end{minipage}
	\begin{minipage}[b]{0.45\linewidth}
	
		\centering
		\subfloat[\label{fig5:b}\footnotesize PrVXM]{
			\includegraphics[height=4.1cm]{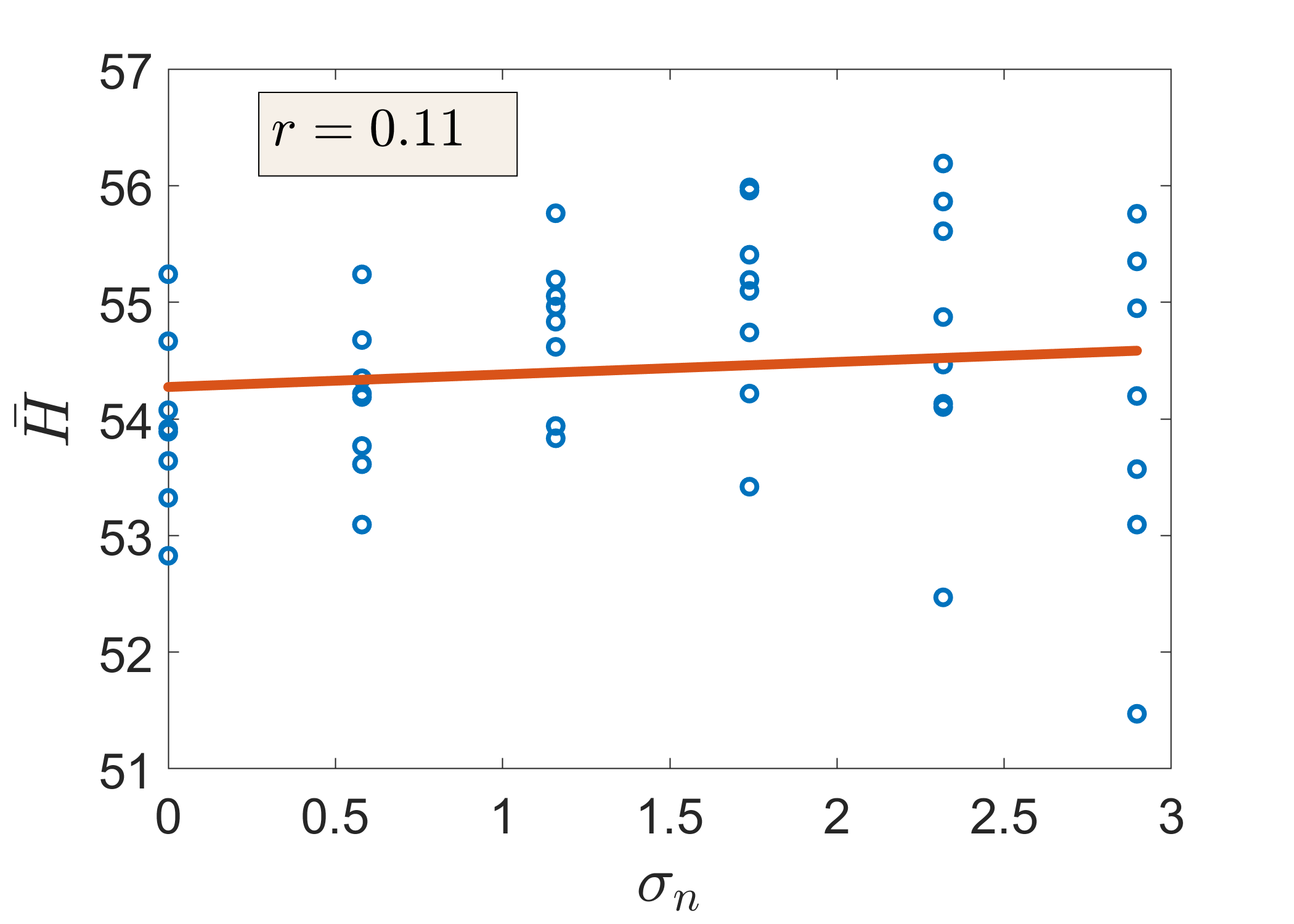}}
	\end{minipage}	
	\caption{Correlation between the NPBDREG and PrVXM estimate of uncertainty and out-of-distribution data. \protect\subref{fig5:a} and \protect\subref{fig5:b} Scatter plots of the mean value of uncertainty estimate, $\Bar{H}$, of NPBDREG and PrVXM versus the amount of noise added to the input images, $\sigma_n$. In contrast with PrVXM, the uncertainty measure of NPBDREG exhibits a high correlation with the noise level ($r>0.95$). $r$ denotes the computed Pearson correlation coefficient.  }\label{fig:unccornoise}	
\end{figure}

 The uncertainty in the velocity field, as mentioned in \eqref{eq:Hcalc}, is a 3D map. An example of slice of the computed voxel-wise uncertainty map is observed in Fig.~\ref{fig:uncexample} (See $H_x$). Fig.~\ref{fig:unccornoise} depicts the correlation between the measure of uncertainty and the level of data corruption by noise. We calculate the mean uncertainty over the overall voxels for each example. We measured the correlation between the two variables by computing the Pearson correlation coefficient, $r$. The high correlation ($r=0.96$, $p<10^{-4}$) indicates the ability of the NPBDREG uncertainty measures to detect out-of-distribution data that would result in unreliable DNN predictions. The out-of-distribution data was generated by adding an increasing amount of noise to the pairs of MRI data serving as the registration DNN-based input.

\section{Conclusions} \label{sec:conc}
In this paper we developed a non-parametric Bayesian DNN-based method to assess the uncertainty in DNN-based diffeomorphic deformable MRI registration.
We used noise injection for the training loss gradients to efficiently sample the true posterior distribution of the network weights. Our proposed approach formulated the training mechanism for an \texttt{Adam} optimizer. however, it can be directly extended to other stochastic optimization algorithms as well.  

The proposed approach provides empirical estimates of the two principle moments of the deformation field, which, in turn, can be directly employed to assess the uncertainty in the deformation field. In addition, other statistics or higher-order moments can be empirically calculated from the posterior samples. Our experiments showed that our non-parametric Bayesian approach for DNN-based registration provides uncertainty measures that are highly correlated with the presence of out of distribution data. Moreover, we demonstrated also an improvement in the registration accuracy as well as a better generalization capability compared to the baseline Probabilistic \texttt{VoxelMorph} model. 

\section*{Acknowledgments}
Khawaled, S. is a fellow of the Ariane de Rothschild
Women Doctoral Program.
\bibliographystyle{model2-names.bst}\biboptions{authoryear}
\bibliography{refs}
\appendix
\section{Convergence Analysis} \label{app:1}
The sampled model parameters, $\theta^{t}$, converge to the posterior distribution when $t\rightarrow\infty$, i.e. $\E\left[\theta^{t}\right]\overset{t\rightarrow\infty}{\longrightarrow}\E\left[\theta\right]$. Therefore, the mean velocity field, calculated by averaging the velocity field samples obtained by the model, $\hat{V}=\tfrac{\sum_{t_b}^{N}w^t V^t}{\sum_{t_b}^{N}w^t}$, is also unbiased when $t\rightarrow\infty$ (consistent). According to SGLD \cite{welling2011bayesian}, convergence to a local minima is guaranteed when the following requirements on the step size are accomplished:
\begin{equation}
\begin{cases}
\begin{array}{c}
\sum_{t=1}^{\infty}\epsilon^{t}=\infty\\
\\
\sum_{t=1}^{\infty}(\epsilon^t)^{2}<\infty
\end{array}\end{cases}
\label{eq:1}
\end{equation}
Typical step sizes decay polynomialy: $\epsilon^{t}=\frac{a}{(b+t)^\gamma}$ when $\gamma\in\left(0.5,1\right]$. In our NPBDREG training, we use the \texttt{Adam} step size, denoted by $s^{t}$, and we select a proper step size that satisfies the SGLD converges. Here we provide a condition on the step size that meets the requirements mentioned in \eqref{eq:1}.  
\paragraph{Converges Condition in NPBDREG scheme}
Let $s^t$ and $\epsilon^t$ denote the \texttt{Adam} and \texttt{SGLD} step sizes, respectively. The former can be written as $s^t=\tfrac{\eta}{\sqrt{\nicefrac{v^{t}}{1-\beta_{2}^{t}}+\epsilon}}$, where  $\beta_{2}^{t}$ is the decay rate, which can be either fixed or varying with training iterations. Here we assume that $\epsilon^t$ is related to the aforementioned polynomialy decaying functions. To fulfill the constraints in \eqref{eq:1}, the following should be accomplished: $s^t\in \Omega\left(\epsilon^t \right)$, i.e. there exist some constants $c_1$ and $c_2$ such that $c_1\frac{a}{(b+t)^\gamma}\leq s^t \leq c_2\frac{a}{(b+t)^\gamma}$. The latter yields to the following two conditions on $\eta$ and $\beta_2^t$: 
 
\begin{equation}
\begin{cases}
\begin{array}{c}
\eta^2 (1-\beta_2^t) \leq v^t c_1^2 \frac{a^2}{(b+t)^{2\gamma}}   \\
\\
\eta^2 (1-\beta_2^t) \geq v^t c_2^2 \frac{a^2}{(b+t)^{2\gamma}}
\end{array}\end{cases}
\label{eq:2}
\end{equation}
In others words, the aforementioned requirements imply that $\beta_{2}^t\in \Omega\left(\frac{v^t}{(b+t)^{2\gamma}}\right)$.

\section{Output Statistics}\label{app:2}
In \cite{cheng2019bayesian}, it is proved that a CNN with random parameters, which is fed by a stationary input image (such as white noise), acts as a spatial Gaussian process with a stationary kernel. This is valid in the limit as the channels number in each layer goes to infinity. In addition, the authors analyze the network statistical behavior for models with beyond two layers or with more complex systems that incorporate down-sampling, sampling or skip connection. Similar to these conclusions, the deformation obtained by a CNN model with random parameters that operates on stationary pair of images behaves like a Gaussian field. We assume that the network parameters, $\theta$ are Gaussian. In our \textit{adaptive} learning setting, the latter is accomplished under the assumption of negligible second order moment of \texttt{Adam}, $v^t$, i.e. when $\beta_{2}\simeq1$.
\section{Anatomical Regions and Implemetation Details} \label{appendix:details}

\begin{table}[ht]
\centering
\scalebox{0.7}{
\begin{tabular}{|l|l|}
\hline
\textbf{Database} & \multicolumn{1}{c|}{\textbf{Labels}}        \\ \hline
MGH10             & \begin{tabular}[c]{@{}l@{}}L temporal pole, R temporal pole,\\ R occipital cortex, L occipital cortex,\\ L frontal pole, R frontal pole\end{tabular}                               \\ \hline
CUMC12            & \begin{tabular}[c]{@{}l@{}}R Cerebral White Matter, L Putamen, \\ L Thalamus Proper, R Putamen, \\ Brain Stem, R Thalamus Proper\end{tabular}                                      \\ \hline
ISBR18            & \begin{tabular}[c]{@{}l@{}}R Frontal pole, R Supracalcarine cortex, \\ L Intracalcarine cortex, R Intracalcarine cortex,\\ L Supracalcarine cortex, L Frontal pole\end{tabular}    \\ \hline
LPBA40            & \begin{tabular}[c]{@{}l@{}}L middle orbitofrontal G, R precentral G, \\ L precentral G, L lateral orbitofrontal G, \\ R middle orbitofrontal G,  R inferior frontal G\end{tabular} \\ \hline
\end{tabular}}
\caption{Anatomical regions used for Dice calculation for each one of the databases. R and L denotes right and left, respectively.}\label{tab:anatomicallabels}
\end{table}

\begin{table}[th]
\centering
\scalebox{0.7}{
\begin{tabular}{|l|l|}
\hline
$N$                      & $4000$                     \\ \hline
$t_b$                  & $3992$                      \\ \hline
learning rate          & $2^{-4}$ \\ \hline
Regularization Weight  & $10^{-7}$                 \\ \hline
Integration Steps ($T$)  & $6$                         \\ \hline
$\lambda$ & $0.1$                      \\ \hline
\end{tabular}
}
\caption{hyperparamters that were used in our experiments.}\label{tab:implementdet}
\end{table}

\begin{table}[ht]
\centering
\scalebox{0.9}{
\begin{tabular}{rrrrr}
\toprule
     $\sigma=0$ &  $\sigma=0.4$ &  $\sigma=0.6$ &  $\alpha=0.2$ &  $\alpha=0.4$ \\
\midrule
3.773109e-07 &  0.095189 &  0.704813 &  0.000014 &  0.000422 \\
\bottomrule
\end{tabular}}\caption{P-values calculated in the paired t-test, which was conducted between Dice scores of the two methods: NPBDREG (ours) and the benchmark PrVXM. $\sigma$ and $\alpha$ denotes the added noise levels in Gaussian and Mixed Structure noise types, respectively.  }\label{tab:pvalues}
\end{table}

\end{document}